\documentclass[journal]{IEEEtran}
\usepackage[pagebackref=true,breaklinks=true,colorlinks,bookmarks=false]{hyperref}
\usepackage{amsmath}
\usepackage{url}
\usepackage{bm}
\usepackage{graphicx}
\usepackage{subfigure}
\usepackage{comment}
\usepackage{amsmath,amssymb} 
\usepackage{color}
\usepackage{multirow}
\usepackage{booktabs}
\usepackage{array}
\usepackage{lineno}
\usepackage{subeqnarray}
\usepackage{cases}
\usepackage[misc]{ifsym}
\usepackage[linesnumbered,boxed,ruled,commentsnumbered]{algorithm2e}
\usepackage{ulem}
\newcommand{\eat}[1]{}

\ifCLASSINFOpdf
\else
\fi
\begin{document}
%
\title{Patch-wise++ Perturbation for Adversarial Targeted Attacks}
%
%
%

\author{Michael~Shell,~\IEEEmembership{Member,~IEEE,}
        John~Doe,~\IEEEmembership{Fellow,~OSA,}
        and~Jane~Doe,~\IEEEmembership{Life~Fellow,~IEEE}
\thanks{M. Shell was with the Department
of Electrical and Computer Engineering, Georgia Institute of Technology, Atlanta,
GA, 30332 USA e-mail: (see http://www.michaelshell.org/contact.html).}
\thanks{J. Doe and J. Doe are with Anonymous University.}
}

\author{Lianli Gao,
	Qilong Zhang,
	Jingkuan Song, 
	and Heng Tao Shen,~\IEEEmembership{Fellow,~ACM}
	\IEEEcompsocitemizethanks{\IEEEcompsocthanksitem Lianli Gao, Qilong Zhang, Jingkuan Song and Heng Tao Shen are with the Future Media Center and School of Computer Science and Engineering, The University of Electronic Science and Technology of China, Chengdu, China, 611731. E-mail: qilong.zhang@std.uestc.edu.cn
	}
}

\maketitle

\begin{abstract}
Although great progress has been made on adversarial attacks for deep neural networks (DNNs), their transferability is still unsatisfactory, especially for targeted attacks. 
There are two problems behind that have been long overlooked: 
1) the conventional setting of $T$ iterations with the step size of $\epsilon/T$ to comply with the $\epsilon$-constraint. In this case, most of the pixels are allowed to add very small noise, much less than $\epsilon$; and 
2) usually manipulating pixel-wise noise. 
However, features of a pixel extracted by DNNs are influenced by its surrounding regions, and different DNNs generally focus on different discriminative regions in recognition.
To tackle these issues, our previous work proposes a patch-wise iterative method (PIM) aimed at crafting adversarial examples with high transferability. Specifically, we introduce an amplification factor to the step size in each iteration, and one pixel's overall gradient overflowing the $\epsilon$-constraint is properly assigned to its surrounding regions by a project kernel. But targeted attacks aim to push the adversarial examples into
the territory of a specific class, and the amplification factor may lead to underfitting. Thus, we introduce the temperature and propose a patch-wise++ iterative method (PIM++) to further improve transferability without significantly sacrificing the performance of the white-box attack.
Our method can be generally integrated to any gradient-based attack methods. 
Compared with the current state-of-the-art attack methods, we significantly improve
the success rate by 33.1\% for defense models and 31.4\% for normally trained models on average. 
\end{abstract}


%
\IEEEpeerreviewmaketitle

\section{Introduction}
\label{Introduce}

%
%
%
%
\IEEEPARstart{W}{ith} the great achievement of deep neural networks (DNNs)~\cite{ref_article29,ref_article30,ref_article28,ref_article27}, various fields have applied them to improve the performance. Therefore, the robustness and stability of DNNs are critical. Unfortunately, recent works have demonstrated that adversarial examples~\cite{ref_article2,Bi+13} which are added with human-imperceptible noise can easily fool the state-of-the-art DNNs to give unreasonable predictions. Due to the vulnerability of DNNs, many researchers have paid close attention to the security problem of these machine learning algorithms. To understand DNNs better and improve their robustness to avoid future risks~\cite{ref_article4}, it is necessary to investigate the generation of adversarial examples.
\begin{figure}[htp]
	\centering
	\includegraphics[height=7cm]{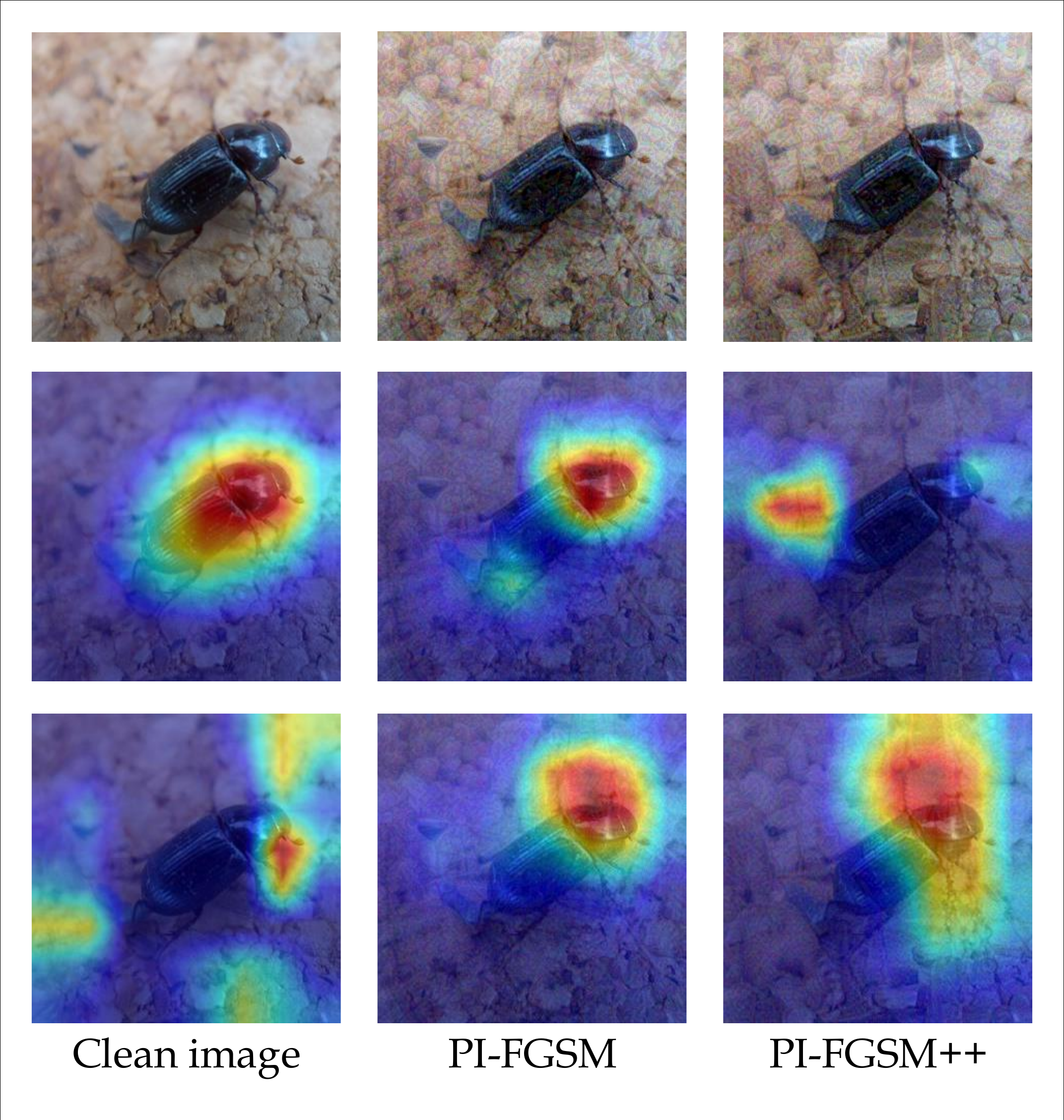}
	\caption{We show the natural image and targeted adversarial examples crafted by our PI-FGSM~\cite{Gao+20} and PI-FGSM++ via an ensemble of Inc-v4, IncRes-v2~\cite{ref_article28}, Res-50, Res-101 and Res-152~\cite{ref_article29} in \textbf{the top row}. Here we use Inc-v3~\cite{ref_article27} which serves as black-box model to show the Gradient-weighted Class Activation Mapping (Grad-CAM)~\cite{ref_article33} of true label ``dung beetle" in \textbf{the middle row} and target label ``scale" in \textbf{the bottom row}. For PI-FGSM, the resultant adversarial example cannot fool Inc-v3 towards ``scale" (confidence $\textless$ 0.02\%). But PI-FGSM++ can simply improve the confidence of ``scale" upon to 99.99\%, and increase the response to the target label.}
	\label{cam}
\end{figure}
In general, attack methods can be divided into three categories according to the available information. The first is the white-box setting (\textit{e.g.},~\cite{deepfool,c-w,curl}) where the adversary can get the full knowledge of the victim's models, thus obtaining accurate gradient information to update adversarial examples. The second is the semi-black-box setting, where only the outputs of the victim's models are available but the specific parameter and structure are still unknown. For example, Papernot \textit{et al.}~\cite{ref_article5} train a local model with many queries to substitute for the target victim's model. Ilyas \textit{et al.}~\cite{ref_article6} propose the variant of NES~\cite{ref_article7} to generate adversarial examples with limited queries. The third is the black-box setting where the adversary generally cannot access the target model and adversarial examples are usually crafted via the substitute model. In this case, the transferability of adversarial examples plays a crucial role. Recently, the black-box attack is a research hotspot and various excellent works~\cite{F,F_E,Li+20,ssm,cheng2021feature} have been proposed. Dong \textit{et al.}~\cite{ref_article10} propose a momentum-based iterative algorithm to boost attack ability. Xie \textit{et al.}~\cite{ref_article11} apply random transformations to the input images at each iteration to improve transferability. Dong \textit{et al.}~\cite{ref_article12} propose translation-invariant attack method~\cite{ref_article12} which can evade the defenses with effect. Lin \textit{et al.}~\cite{nisim} adapt Nesterov accelerated gradient and leverage scale-invariant property of DNNs to optimize the perturbations. Wu \textit{et al.}~\cite{sgm} explore the security weakness of skip connections~\cite{ref_article29} to boost adversarial attacks.

To generate more transferable adversarial examples, it is necessary to study the properties of the intrinsic classification logic of the DNNs. In our previous work~\cite{Gao+20}, we adopt {class activation mapping} \cite{ref_article34} to analyze it. Inspired by several related works~\cite{ref_article34,ref_article12}, we summarize three characteristics of the discriminant regions: 1) The discriminative regions always vary across predicted labels. 2) Different models generally focus on different discriminative regions, and the defense models generally focus on larger discriminative regions than the normally trained models. 3) Besides, the discriminative regions are often clustered together, \textit{i.e.}, highlighted regions always merge together (see Fig.~\ref{cam}). Therefore, only adding pixel-wise noise through small step size may hinder the transferability of adversarial examples across different DNNs. 
Motivated by it, we argue that patch-wise (\textit{i.e.}, regionally homogeneous) noise  will be more effective than the pixel-wise one,
by which varied discriminative regions of different DNNs can be better covered.

\eat{
To the best of our knowledge, most of existing works unnecessarily set $T$ iterations with step size of $\epsilon/T$, \textit{e.g.},~\cite{Li+20,ref_article10,ref_article12}, to comply with the $\epsilon$-constraint. By this way, although the $L_\infty$-norm of the resultant noise can finally reach the maximum perturbation $\epsilon$, most of the pixels are allowed to add very small noise, and thus adversarial examples are more likely be captured by poor local optimum. 
Based on the aforementioned motivations, we propose a broad class of patch-wise algorithm to boost the transferability of adversarial examples.
}

To craft effective patch-wise noise, we review the advantages and disadvantages of single-step and iterative attacks \cite{ref_article12,ref_article8,ref_article9}, and argue that linear nature of DNNs \cite{ref_article8} does exist to some extent. Thus, we amplify the step size with a fixed factor (\textit{i.e.}, the amplification factor) to increase the average magnitude of the noise. Besides, we rethink the weakness of direct clipping operation which discards partial gradient information. To alleviate this problem, we apply a heuristic project strategy to reduce the side effects of direct clipping. Combined the amplification factor and the heuristic project strategy, we propose a non-targeted \textbf{Patch-wise Iterative Method (PIM)}~\cite{Gao+20} to boost adversarial attack.

However, targeted attack (\textit{i.e.}, fooling the victim's models to give pre-set untrue labels with high confidence) is more challenging than non-targeted attacks. In general, targeted attacks need an ensemble of substitute models to achieve an acceptable attack success rate. Although Li~\textit{et al.}~\cite{Li+20} have increased the targeted attack success rate by introducing $Poincar\acute{e}$ distance as a similar metric to alleviate noise curing and triple loss to push update direction away from the true label, the resultant adversarial examples are still challenging to transfer to black-box models, especially for defense models.

To tackle this issue, in this paper, we boost targeted attacks based on our PIM. Considering that the white-box model is usually an ensemble of models, the resultant adversarial examples are born with high transferability. As a result, simply applying the amplification factor of PIM may induce the update direction unstable.  
Therefore, we propose targeted \textbf{Patch-wise++ Iterative Method (PIM++)}, in which we apply the temperature term to soften the output probability distribution at each iteration to alleviate the underfitting problem which may be caused by the amplificaiton factor. In this way, resultant adversarial examples can better approximate the distribution of the natural target images.
Similar to PIM~\cite{Gao+20}, our PIM++ can be generally integrated to any gradient-based attack methods, \textit{\textit{e.g.}}, fast sign gradient method (FGSM)~\cite{ref_article8}. 
Compared with state-of-the-art methods, our PIM++ boosts targeted attack ability by a large margin.

To sum up, our major contributions can be summarized as:
\textbf{1)} We first propose a novel patch-wise iterative method named PIM for the non-targeted attack. Specifically, we adopt an amplification factor and a project kernel to generate more transferable adversarial examples.
\textbf{2)} Considering that the target attacks need to push the adversarial examples into a specific territory of a target class rather than out of the original class's region, the amplification factor may lead to underfitting. Therefore, we propose PIM++ for target attacks, in which we soften the output probability distribution at each iteration to alleviate the underfitting problem. Our approach can have the advantages of both single-step and iterative attacks, \textit{i.e.}, improving the transferability without sacrificing the performance of the substitute model.
\textbf{3)} Extensive experiments on ImageNet show that our targeted method significantly outperforms the state-of-the-art methods, and improves the success rate by \textbf{33.1\%} for defense models and \textbf{31.4\%} for normally trained models on average in the black-box setting. 
The source code for PIM is available at \url{https://github.com/qilong-zhang/Patch-wise-iterative-attack} and the source code for PIM++ is available at \url{https://github.com/qilong-zhang/Targeted_Patch-wise-plusplus_iterative_attack}.

The remainder of this paper is organized as follows, we first briefly review the related work in Sec.~\ref{related work}, and illustrate the process of generating adversarial examples in Sec.~\ref{{preliminaries}}. 
Then, we extend PIM to PIM++ in Sec.~\ref{pim++}. Finally, we conduct extensive experiments to compare the results of the proposed method with other state-of-the-art methods in Sec.~\ref{exp}.



\section{Related work}
\label{related work}
In this section, we first introduce what are adversarial examples in Sec.~\ref{ae}, then briefly list several defense methods in Sec.~\ref{dm}, and finally discuss the ensemble learning in Sec.~\ref{el}.

\subsection{Adversarial Examples}
\label{ae}
Adversarial examples~\cite{Bi+13,ref_article2,che2021adversarial} are a particularly worrisome phenomenon, which only add subtle perturbation to the clean images but can mislead the DNNs to make an unreasonable prediction with unbelievably high confidence. To make matters worse, adversarial examples also exist in physical world \cite{ref_article4,ref_article19,ref_article3}, which raises security concerns about DNNs. Due to the vulnerability of DNNs, a large number of attack methods have been proposed and applied to various fields of deep learning in recent years, \textit{e.g.}, object detection and semantic segmentation \cite{DBLP:conf/iccv/XieWZZXY17}, embodied agents \cite{Liu2020Spatiotemporal}, and speech recognition \cite{DBLP:journals/corr/CisseANK17}. To make our paper more focused, we only analyze adversarial examples in the image classification task.

\subsection{Defense Method}
\label{dm}
With the great achievement of attack methods, several adversarial examples have been successfully applied to the physical world~\cite{Liu2020Biasbased,ref_article4,ref_article19,ref_article17,ref_article18}. And this technique has also raised public concerns about AI security. Consequently, a lot of defense methods are proposed to tackle this problem. Guo \textit{et al.}~\cite{ref_article21} use bit-depth reduction, JPEG compression~\cite{ref_article22}, total variance minimization~\cite{ref_article23} and image quilting~\cite{ref_article24} to preprocess inputs before they are feed to DNNs. Tramèr \textit{et al.}~\cite{ref_article25} use {ensemble adversarial training} to improve the robustness of models. Furthermore, Xie \textit{et al.}~\cite{ref_article26} add feature denoising module into adversarial training.
Mustafa \textit{et al.}~\cite{mustafa2020image} propose an image restoration scheme based on super-resolution to mitigate the effect of adversarial perturbations.	
Zhang~\textit{et al.}~\cite{zhang2021interpreting} stable the behaviors of sensitive neurons to improve the robustness of the model.

\subsection{Ensemble Learning}
\label{el}

Ensemble learning has been applied in researches and competitions to improve performance. This idea is so pervasive that it can be easily adopted in the adversarial attack. Liu \textit{et al.}~\cite{ref_article16} first utilize an ensemble of multiple models' predictions to generate adversarial examples, and they demonstrate ensemble-based approaches can overcome the problem of low targeted transferability because the adversarial examples are less likely to get stuck in the local optimum of any specific models. Further, Dong \textit{et al.}~\cite{ref_article10} fuse the logits instead of predictions or losses, and the experimental results demonstrate that this fusion strategy is better than others. In this paper, we also fuse the logits when attacking an ensemble of $K$ models:
\begin{equation}
	l(\bm{x}) = \sum_{k=1}^K w_kl_k(\bm{x})
	\label{ensemble}
\end{equation}
where $l_k(x)$ denotes the logits (before the \textit{softmax} function) of $k$-th model and $w_k$ is the weight for each model with $w_k\geq 0$ and $\sum_{k=1}^K w_k = 1$. 


\section{Preliminaries}
\label{{preliminaries}}
In this section, we describe the process of generating adversarial examples in detail. Let $\bm{x}$ denote a clean example without any perturbation and $y$ denote the corresponding true label. We use $f(\bm{x})$ to denote the prediction label of the classifier, and $\bm{x^{noise}}$ to denote the human-imperceptible perturbation. The adversarial example $\bm{x^{adv}} = \bm{x} + \bm{x^{noise}}$ is visually indistinguishable from $\bm{x}$ but misleads the classifier to give high confidence of a wrong label. 
In this paper, we focus on \textbf{targeted transfer-based black-box attacks}, \textit{i.e.}, $f(\bm{x^{adv}}) = y^{adv}$, where $y^{adv}$ is pre-set target label and $y^{adv}\neq y$. To measure the perceptibility of adversarial perturbations, we follow previous works \cite{ref_article10,ref_article12,ref_article11}  and apply $l_{\infty}$-norm here. Namely, we set the maximum adversarial perturbation $\epsilon$ and we should keep $||\bm{x}-\bm{x^{adv}}||_{\infty} \leq \epsilon$. To generate our adversarial examples, we should minimize the loss function $J(\bm{x^{adv}}, y^{adv})$. Here $J(\cdot)$ is cross-entropy loss. Therefore, our goal is to solve the following constrained optimization problem:
\begin{equation}
	\underset{\bm{x^{adv}}}{\arg \min} J(\bm{x^{adv}}, y^{adv}), \qquad  s.t.\ ||\bm{x}-\bm{x^{adv}}||_\infty \leq \epsilon.
\end{equation}
Due to the black-box setting, the adversaries do not allow to analytically compute the gradient on the target model. In the majority of cases, they utilize the information of substitute models (\textit{i.e.}, official pre-trained models which serve as white-box models) to generate adversarial examples. But the resultant adversarial examples are highly correlated with the white-box models, which may induce overfitting and hinder the black-box attack success rate.
Therefore, it is very important to improve the transferability of adversarial examples so that they still fool the black-box models successfully.

\subsection{{Development of Gradient-based Attack Methods}}
In this section, we give a brief introduction of some excellent black-box works which are based on the transferability of adversarial examples.

\textbf{Fast Gradient Sign Method (FGSM)}: Goodfellow \textit{et al.}~\cite{ref_article8} argue that the vulnerability of DNN is their linear nature. Consequently they update the adversarial example by: 
\begin{equation}
	\bm{x^{adv}} = \bm{x} - \epsilon \cdot sign(\nabla_{\bm{x}} J(\bm{x}, y^{adv})),
\end{equation}
where $sign(\cdot)$ indicates the sign operation.

\textbf{Iterative Fast Gradient Sign Method (I-FGSM)}: Kurakin \textit{et al.}~\cite{ref_article9} adopt a small step size $\alpha$ to iteratively apply the gradient sign method multiple times. This method can be written as:
\begin{equation} 
	\bm{x^{adv}_{t+1}} = Clip_{\bm{x}, \epsilon}\{\bm{x^{adv}_{t}} - \alpha \cdot sign(\nabla_{\bm{x}} J(\bm{x^{adv}_t}, y^{adv}))\},
	\label{I-FGSM}
\end{equation} 
where $Clip_{\bm{x}, \epsilon}$ denotes element-wise clipping, aiming to restrict $\bm{x^{adv}}$ within the $l_\infty$-bound of $\bm{x}$. 

\textbf{Momentum Iterative Fast Gradient Sign Method (MI-FGSM)}: Dong \textit{et al.}~\cite{ref_article10} apply momentum term to stabilize update directions. It can be expressed as:
\begin{equation}
	\begin{split}
		\bm{g_{t+1}} = \mu \cdot \bm{g_t} + \frac{\nabla_{\bm{x}} J(\bm{x^{adv}_t}, y^{adv})}{||\nabla_{\bm{x}} J(\bm{x^{adv}_t}, y^{adv}) ||_1},~~~~\\
		\bm{x^{adv}_{t+1}} = Clip_{\bm{x},\epsilon}\{\bm{x^{adv}_t} - \alpha \cdot sign(\bm{g_{t+1}})\},
		\label{eq.mifgsm}
	\end{split}
\end{equation}
where $\bm{g_t}$ is cumulative gradient, and $\mu$ is the decay factor.

\textbf{Diverse Input Iterative Fast Gradient Sign Method (DI$^2$-FGSM)}: Xie \textit{et al.}~\cite{ref_article11} apply diverse input patterns to improve the transferability of adversarial examples. With the replacement of Eq. (\ref{I-FGSM}) by:
\begin{equation}
	\bm{x^{adv}_{t+1}} = Clip_{\bm{x}, \epsilon}\{\bm{x^{adv}_{t}} - \alpha \cdot sign(\nabla_{\bm{x}} J(D(\bm{x^{adv}_t}), y^{adv}))\},
\end{equation}
where $D(\bm{x})$ is random transformations to the input $x$. For simplicity, we use DI-FGSM later.

\textbf{Translation-Invariant Fast Gradient Sign Method (TI-FGSM)}: Dong \textit{et al.}~\cite{ref_article12} convolve the gradient with the pre-defined kernel $\bm{W}$ to generate adversarial examples which are less sensitive to the discriminative regions of the substitute model. It is only updated in one step:
\begin{equation}
	\bm{x^{adv}} = \bm{x} - \epsilon \cdot sign(\bm{W} *\nabla_{\bm{x}} J( \bm{x^{adv}_t}, y^{adv})),
\end{equation}
and TI-BIM is its iterative version.

\textbf{Poincar$\acute{e}$ Iterative Fast Gradient Sign Method (Po-FGSM)}: Li \textit{et al.}~\cite{Li+20} regularize the targeted attack process with Poincar$\acute{e}$ distance ($J_{Po}(\cdot)$) and Triplet loss ($J_{trip}(\cdot)$) to get overall loss function:
\begin{equation}
	J^*(\bm{x_t^{adv}},y)=J_{Po}(\bm{x_t^{adv}},y) + \lambda \cdot J_{trip}(y^{adv},\bm{x_t^{adv}},y),
\end{equation}
then they guide adversarial examples by above loss:
\begin{equation}
	\bm{x^{adv}_{t+1}} = Clip_{\bm{x}, \epsilon}\{\bm{x^{adv}_{t}} - \alpha \cdot sign(\nabla_{\bm{x}} J^*(\bm{x^{adv}_t}, y^{adv}))\}.
\end{equation}

\textbf{Patch-wise Iterative Fast Gradient Sign Method (PI-FGSM)}: Gao \textit{et al.}~\cite{Gao+20} propose patch-wise perturbation by amplifying the step size and projecting the cut noise in each iteration:
\begin{equation}
	\begin{split}
		\bm{x^{adv}_{t+1}} =\! Clip_{\bm{x}, \epsilon}\{ \bm{x^{adv}_{t}} - \beta \cdot \frac{\epsilon}{T}\cdot sign(\nabla_{\bm{x}} J( \bm{x^{adv}_t}, y)) \\- \gamma \cdot sign(\bm{W_p}*\bm{C})\},
	\end{split}
\end{equation}
where $\beta$ is the amplification factor, $\bm{W_p}$ is the uniform project kernel whose size is $k_w\times k_w$ and $\bm{C}$ is the cut noise.

\section{Patch-wise++ Iterative Method}
\label{pim++}
In our previous work~\cite{Gao+20}, we only focus on non-targeted attacks of patch-wise iterative method (PIM). Although we have achieved great non-targeted transferability, the targeted transferability of PIM remains unknown. Besides, due to the fact that targeted adversarial examples are always crafted via an ensemble of models, most previous works~\cite{ref_article10,Li+20} only consider the performance on the ensemble models (white-box setting) and the hold-out model (black-box setting). However, only these two evaluation metrics may not provide a comprehensive comparison of the performance of different attack methods. Therefore, in this section, we will first introduce a new metric to evaluate the performance of attack methods in Sec.~\ref{aoe}, then rethink the effect of our amplification factor for the targeted attack in Sec.~\ref{rethink}, 
and finally we propose our PIM++ in Sec.~\ref{soft}.
\begin{figure*}
	\centering
	\includegraphics[height=7.5cm]{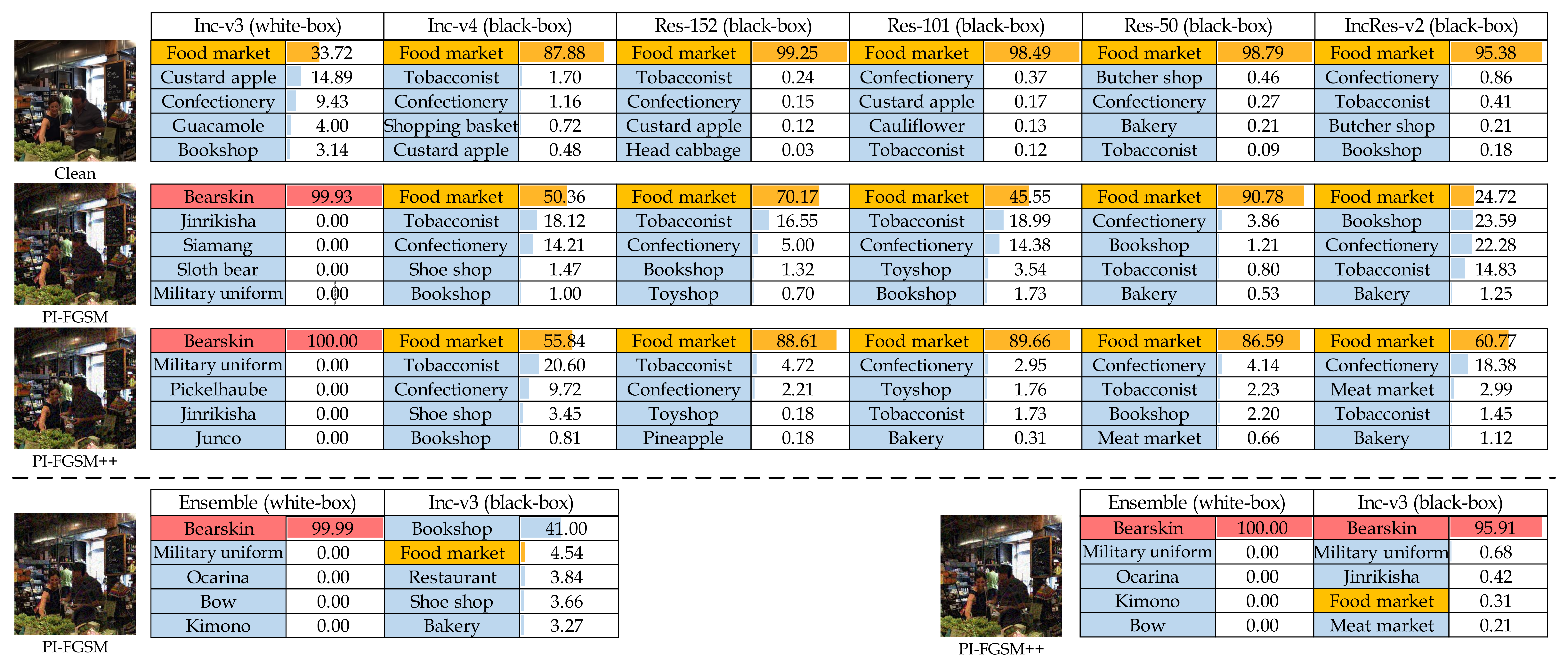}
	\caption{The top-5 confidence of six state-of-the-art DNNs w.r.t left natural images and adversarial examples.
		The true labels and target labels are marked as orange and red respectively. The adversarial examples are crafted by PI-FGSM~\cite{Gao+20} and our proposed PI-FGSM++ with the maximum perturbation $\epsilon=16$. \textbf{The above three rows}: the clean image and its adversarial examples crafted via Inc-v3~\cite{ref_article27}. 
		A first glance shows that natural images are always correctly classified, however, targeted adversarial examples can only fool Inc-v3 (white-box model) but cannot cause the target labels to appear in top-5 confidence of the black-box models. Besides, PI-FGSM++ cannot outperform PI-FGSM in this case. \textbf{The bottom row}: the adversarial examples are crafted via an ensemble of models, \textit{i.e.}, Inc-v4, IncRes-v2~\cite{ref_article28}, Res-50, Res-101 and Res-152~\cite{ref_article29}. Compared with PI-FGSM, PI-FGSM++ is more effective in both black-box and white-box settings.}
	\label{cmp}
\end{figure*}

\subsection{Average of Ensemble}
\label{aoe}
In the previous works~\cite{ref_article10,Li+20}, the evaluation metrics only include the attack success rate of ensemble models (white-box) and hold-out model (black-box) when adversarial examples are crafted via an ensemble of models. However, we observe that the high attack success rate of ensemble models is not necessarily indicative of the high performance of each model. To better evaluate the attack ability of different attack methods, we propose another white-box metric called \textbf{Average of Ensemble (AoE)}. More specifically, our AoE averages the performance among all white-box models:
\begin{equation}
	AoE = \frac{\sum_{k=1}^K S(f_k(\bm{X^{adv}}), Y^{adv})}{K},
\end{equation}
where $\bm{X^{adv}}$ and $Y^{adv}$ denote the set of all resultant adversarial examples and their corresponding target labels, $f_k(\cdot)$ means the output label of $k-$th model, and $S(\cdot, \cdot)$ calculates the targeted success rate. 

With this evaluation metric, we can better measure whether the adversarial examples are in the global optimum (\textit{i.e.}, a high attack success rate of AoE) rather than the local optimum of the ensemble models (\textit{i.e.}, a high attack success rate of ensemble models but the low attack success rate of AoE). 

\subsection{Rethink the Amplification Factor $\beta$}
\label{rethink}
\begin{figure*}
	\centering
	\includegraphics[height=5.3cm]{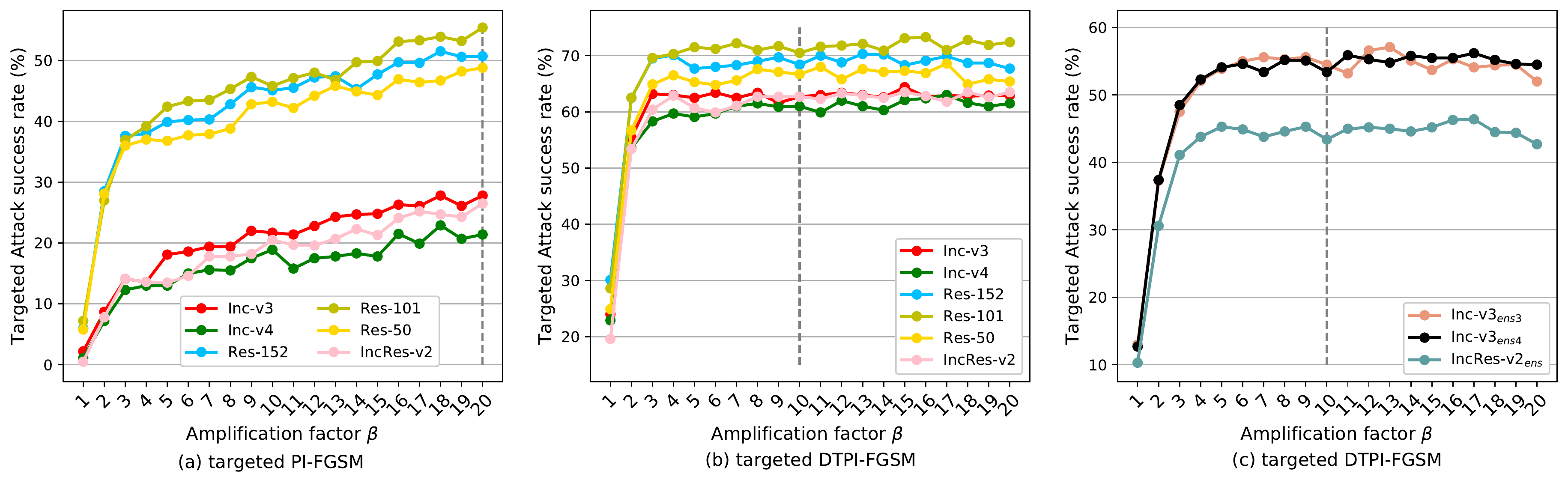}
	\caption{The targeted attack success rate(\%) of each hold-out black-box model w.r.t amplification factor $\beta$. \textbf{(b)}: The adversarial examples are crafted via an ensemble of the remaining five NT by PI-FGSM. \textbf{(a)}: The adversarial examples are crafted via an ensemble of the remaining five NT by DTPI-FGSM. \textbf{(c)}: The adversarial examples are crafted via an ensemble of the remaining six NT and two EAT by DTPI-FGSM.}
	\label{beta}
\end{figure*}

Our previous work PIM introduces a big amplification factor to boost the non-targeted adversarial transferability. In that regard, the goal is to make the resultant classification different from the true label and the amplification factor can certainly avoid falling into local optimum and add more noise to each pixel. 
However, whether a big step can improve the targeted transferability is still unknown. 
Therefore, we conduct experiments in Sec.~\ref{exp}. Surprisingly, as shown in Fig.~\ref{beta}, no matter what the black-box model is, the amplification factor consistently improves the success rates of targeted attacks by a large margin.
One possible reason is that the amplification factor allows pixels to add more noises on average, thus escaping from the local optimum and being close to target distribution.
As demonstrated in the right figure of Fig.~\ref{ablation}, iterative approaches with a suitable amplification factor (\textit{e.g.}, $\beta=4$) will help to increase the success rate of AoE, which supports our above assumption and proves that setting the step size to $\epsilon/T$ is unreasonable.

Although our experiment demonstrates the effectiveness of the amplification factor, the update iteration may be unstable.
Considering that targeted attacks need to push adversarial examples to the specific territory of a target class where most DNNs can misclassify, attack methods with the amplification factor may have a risk of jumping out of this target region, thus leading to underfitting and performing an unsuccessful targeted attack. 
As demonstrated in Tab.~\ref{NT}, single-step targeted attacks with the step size of maximum perturbation $\epsilon$, \textit{e.g.}, FGSM and TI-FGSM, are much more ineffective than their iterative ones with smaller step size, \textit{e.g.}, I-FGSM and TI-BIM. 
While decreasing the step size is a possible solution to alleviate this problem, it inevitably increases the number of iterations, which is inefficient.





\subsection{Our Method: Targeted Patch-wise++ Iterative Method}
\label{soft}
In this paper, we tackle this issue from another perspective. In general, natural images are correctly classified by different DNNs while adversarial examples can not fool them with the same label.
Since different DNNs have distinct decision boundaries for a specific class, natural images are more likely mapped into the intersection of distinct decision regions of them while the adversarial examples are not. Here we call this intersectional region as ``global optimal region''. 
Definitely, we cannot enumerate all the DNNs to find ``global optimal region'' of each class, and push the adversarial examples to their corresponding regions. 
Instead, we use an ensemble of models as our white-box model to estimate these ``global optimal regions''.
{However, due to the side-effect of the amplification factor, it is very possible to craft adversarial examples that are out of the ``global optimal region''.} 
Besides, there is a limitation in \textit{softmax} function when adversarial examples are crafted via an ensemble of models.
{For example, suppose our substitute model is an ensemble of 5 binary models and each model has equal weight. 
For the target label $y^{adv}$, the output logit of an adversarial example is $5$ (the logits of each model is $(20,2,1,1,1)$). For the true label $y$, the output logit of it is $3$ (the logits of each model is $(3,3,3,3,3)$).
Then, after the \textit{softmax} function, it has the probability of $0.88$ for `scale', and $0.12$ for `dung beetle'. Obviously, this adversarial example is out of the ``global optimal region'', but the cross entropy loss will impose a minor penalty on this example.}

To tackle the above issues, we imitate the idea from Hinton \textit{et al.}~\cite{Hinton+15} who first soften output logits of an ensemble of models (also called ``teacher model") to produce a softer probability distribution over classes, and then lightweight ``student model" can learn more useful information from it. In the above case, soft targets can be regarded as regularizers. 



However, in the adversarial targeted attack case, 
we need to perturb the clean images so that they can be misclassified as the target label.
Generally, the update direction is mainly based on the pre-set target label and the substitute models because we cannot access the probability distribution of target clean images to teach adversarial examples.
Thereby, ``teacher" is the target label and the output probability distribution of the substitute models is more like ``student" in our case. 
Although this is different from \cite{Hinton+15}, we can still take the idea from the soft output inspired by~\cite{ziqi,goldblum}. 
More specifically, we divide the output logits of adversarial examples by temperature $\tau$ before being fed into the \textit{softmax} function at each iteration:
\begin{equation}
	\label{te}
	l'(\bm{x})=l(\bm{x}) / \tau,
\end{equation}
where $l(\bm{x})$ are the output logits of an ensemble of white-box models. Any temperature $\tau$ greater than 1 will soften the probability distribution after the \textit{softmax} function. Then, a larger penalty will be imposed even if the original cross-entropy loss of the adversarial example is already small. For example, if $\tau=10$, the adversarial example mentioned above will have the probability of $0.55$ for `scale' and $0.45$ for `dung beetle'. In contrast to the amplification factor, this operation can be regarded as an overfitting process. As a result, the adversarial example is more likely to be pushed into the ``global optimal region".

 


By adding Eq.~\ref{te} into the update process, our patch-wise perturbation upgrades to patch-wise++ perturbation. Nevertheless, there is also a limitation in this method.
As shown in Fig~\ref{cmp}, when the adversarial examples are generated via a single substitute model, \textit{i.e.}, second and third row, PI-FGSM++ improves the confidence of target label on the substitute model Inc-v3 (from 99.93\% to 100.00\%) at the cost of higher confidence of original true label on other black-box models. For example, the confidence of ``Food market" on Res-101 increases by 44.11\%, which is useless for improving the transferability. This is because the decision boundaries of different models are generally distinct and overfitting on one model cannot generalize well to other black-box models. But with the help of an ensemble of models, the global decision boundary can be better estimated and the limitation can also be ignored. As demonstrated in the bottom row of Fig.~\ref{cmp}, the adversarial example crafted by PI-FGSM++ can successfully mislead the black-box model to give high confidence of target label. 
\IncMargin{1em} 
\begin{algorithm*}[t]
	\DontPrintSemicolon
	\SetAlgoNoLine 
	\SetKwInOut{Input}{\textbf{Input}}\SetKwInOut{Output}{\textbf{Output}}
	\Input{The cross-entropy loss function $J$ of our substitute models; logits before the \textit{softmax} operation of $k-$th model $l_k(\cdot)$; temperature $\tau$;iterations $T$; $L_{\infty}$ constraint $\epsilon$; project kernel $\bm{W_p}$; amplification factor $\beta\,(\geq 1)$; project factor $\gamma$; a clean image $\bm{x}$ (Normalized to [-1,1]) and the target label $y^{adv}$; 
	}
	\Output{The adversarial example $\bm{x^{adv}}$;
	}
	Initialize cumulative amplification noise $\bm{a_0}$ and cut noise $\bm{C}$ to 0;\quad\\
	$\bm{x^{adv}_0} = \bm{x}$;
	
	\For{$t \leftarrow 0$ \KwTo $T$}
	{
		$l(\bm{x^{adv}_t}) = \sum_{k=1}^K w_kl_k(\bm{x^{adv}_t})$;\quad\\
		$l'(\bm{x^{adv}_t})=l(\bm{x^{adv}_t}) / \tau$;\quad\\
		Calculate the gradient $\nabla_{\bm{x}} J( \bm{x^{adv}_t}, y^{adv})$ based on $l'(\bm{x^{adv}_t})$ rather than $l(\bm{x^{adv}_t})$;\quad\\
		
		$\bm{a_{t+1}} = \bm{a_{t}} + \beta \cdot \frac{\epsilon}{T} \cdot sign(\nabla_{\bm{x}} J(\bm{x^{adv}_t}, y^{adv}))$; \tcp*{Update $\bm{a_{t+1}}$}
		\eIf{$||\bm{a_{t+1}}||_{\infty} \ge \epsilon$} 
		{	
			$\bm{C} = clip(|\bm{a_{t+1}}|-\epsilon, 0, \infty) \odot sign(\bm{a_{t+1}})$;\quad\\
		}
		{
			$\bm{C} = 0$;
		}
		$\bm{x^{adv}_{t+1}} = Clip_{\bm{x}, \epsilon}\{\bm{x^{adv}_{t}} - \beta \cdot \frac{\epsilon}{T} \cdot sign(\nabla_{\bm{x}} J(\bm{x^{adv}_t},  y^{adv})) - \gamma \cdot sign(\bm{W_p}*\bm{C})\}$; 	\quad\\
		$\bm{x^{adv}_{t+1}} = clip(\bm{x^{adv}_{t+1}}, -1, 1)$; \tcp*{Finally clip $\bm{x^{adv}_{t+1}}$ into [-1,1]}
	}
	Return $\bm{x^{adv}} = \bm{x^{adv}_T}$;
	\caption{Targeted PI-FGSM++\label{A2}}
\end{algorithm*}
\DecMargin{1em}

To make our conclusion more convincing, we apply the Grad-CAM~\cite{ref_article33} to analyze it in Fig.~\ref{cam}. In the middle row, we show the true label's results. A first glance shows that PI-FGSM cannot completely remove the positive response on the ``head" of ``dung beetle" but PI-FGSM++ does it, which demonstrates that the resultant adversarial examples are further away from the true label. In the bottom row, we show the target labels' results. 
The visualization demonstrates that our PI-FGSM++ 
can generate more effective adversarial noise than PI-FGSM to increase the response to the target label, \textit{i.e.}, larger discriminative region. 

Our method (PI-FGSM++) is summarized in Algorithm.~\ref{A2}. First, we apply temperature $\tau$ to soften the probability distribution of the substitute model, \textit{i.e.}, an ensemble of models. Then, we amplify the step size by $\beta$ at each iteration and calculate the cumulative amplification noise $\bm{a_t}$
If $L_\infty$-norm of $\bm{a_t}$ exceeds the threshold $\epsilon$, we update the cut noise $\bm{C}$ by:

\begin{equation}
	\bm{C} = clip(|\bm{a_{t+1}}|-\epsilon, 0, \infty) \odot sign(\bm{a_{t+1}}),
\end{equation}
where $|\cdot|$ denotes the absolute operation. 
After that, we can add an additional project term $\gamma \cdot sign(\bm{W_p}*\bm{C})$ to the perturbation. 
Unlike non-targeted attacks, we apply gradient descent rather than gradient ascent at each iteration to craft adversarial examples:
\begin{equation}
	\begin{split}
		\bm{x^{adv}_{t+1}} \!\!=\!\! Clip_{\bm{x}, \epsilon}\{\bm{x^{adv}_{t}} \!\!-\!\! \beta \!\cdot\!\! \frac{\epsilon}{T} \!\!\cdot\! sign(\nabla_{\bm{x}} J(\bm{x^{adv}_t},\! y^{adv})) \\ - \gamma \cdot sign(\bm{W_p}*\bm{C})\},
	\end{split}
\end{equation}
where $\bm{W_p}$ is a uniform project kernel whose size is $k_w\times k_w$:
\begin{equation}
	\label{W_p}
	\bm{W_p[i,j]} =  \begin{cases}
		0 ,&  i = \lfloor k_w/2 \rfloor, j =\lfloor k_w/2 \rfloor. \\
		
		1 / (k_w^2-1) ,& else.
	\end{cases}
\end{equation}
Finally we clip the adversarial examples $\bm{x^{adv}}$ into $\epsilon$-ball of $\bm{x}$ and restrict them within [-1,1]. As demonstrated in Fig.~\ref{cam} and Fig.~\ref{cmp}, our PI-FGSM++ is more effective than its baseline PI-FGSM.

\section{Experiments}
\label{exp}

In this section, we present the experimental results to demonstrate the effectiveness of our proposed method. We first define the Setup in Sec.~\ref{setup}.
Considering that our PIM++ is based on PIM, we first extend the non-targeted PI-FGSM to a targeted version, and then discuss the optimal parameters, \textit{i.e.}, amplification factor $\beta$, project factor $\gamma$ and project kernel length $k_w$, in Sec.~\ref{tpw}.
Following the same parameter settings, we further analyze the effectiveness of temperature $\tau$ in Sec.~\ref{ttpn}. Finally, we compare the performance of our targeted patch-wise and patch-wise++ iterative methods with other state-of-the-art methods in Sec.~\ref{attacking nt} and Sec.~\ref{attacking eat} to respectively demonstrate the targeted transferability on NT and EAT.
\subsection{Setup}
\label{setup}
\textbf{Networks}. To avoid cheery-picking, here we follow the recent work~\cite{Li+20} and consider six state-of-the-art normally trained models: Inc-v3~\cite{ref_article27}, Inc-v4, IncRes-v2~\cite{ref_article28}, Res-50, Res-101 and Res-152~\cite{ref_article29} and three ensemble adversarial trained defense models: Inc-v3$_{ens3}$, Inc-v3$_{ens4}$ and IncRes-v2$_{ens}$ \cite{ref_article25} as our victim's models.

\textbf{Dataset}. Following the previous works \cite{Gao+20}, we also conduct our experiments on ImageNet-compatible dataset\footnote{\url{https://github.com/tensorflow/cleverhans/tree/master/examples/nips17_adversarial_competition/dataset}}, which contains 1,000 images and is used for NIPS 2017 adversarial competition. 

\textbf{Abbreviation}. Since PIM and PIM++ can be easily combined with other attack methods (\textit{e.g.}, DI-FGSM~\cite{ref_article11}). To make the abbreviation unambiguous, we use the first character to denote the corresponding method. For instance, DPI-FGSM++ means the integration of DI-FGSM with PIM++. Besides, we use NT to denote normally trained model(s), EAT to denote ensemble adversarial trained model(s). Besides, for methods whose step size is amplified by our amplification factor $\beta$, we use the character ``A" to denote them, \textit{e.g.}, AI-FGSM~\cite{Gao+20} denotes the resultant method that amplifies the step size ($\epsilon / T$) of I-FGSM.

\textbf{Parameter}. In our experiment, we set equal weight when attacking an ensemble of models, \textit{i.e.}, $1/K$ where $K$ is the total number of ensemble white-box models. The maximum perturbation $\epsilon$ is set to 16. The iteration $T$ is set to 20 for all iterative methods. For iterative methods without our amplification factor, the step size $\alpha$ is $\epsilon / T = 0.8$. For MI-FGSM, we set the decay factor $\mu = 1.0$. For TI-FGSM and TI-BIM, we set the kernel size $k = 5$ when the victim's models are NT and $k=15$ when the victim's models are EAT. For DI-FGSM, we set the transformation probability $p=0.7$. For DTMPo-FGSM, we set $\lambda=0.01$ to balance the Poincar$\acute{e}$ distance and triplet loss.
\begin{figure*}
	\centering
	\includegraphics[height=5.3cm]{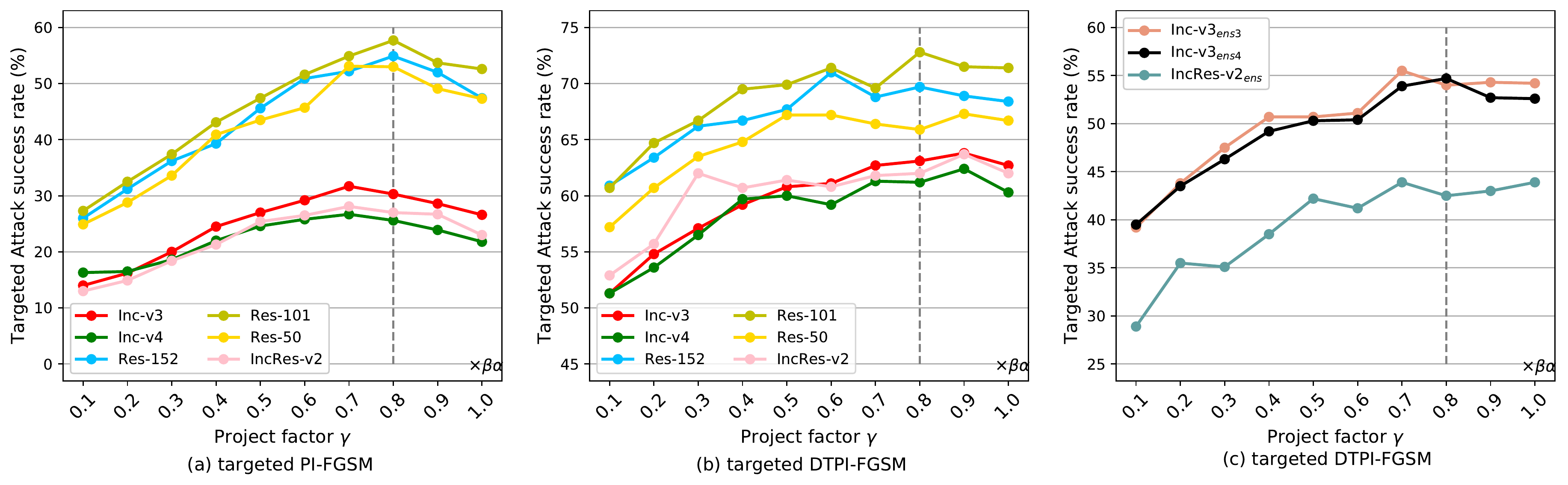}
	\caption{The targeted attack success rate(\%) of each hold-out black-box model w.r.t project factor $\gamma$. \textbf{(a)}: the adversarial examples are crafted via an ensemble of the remaining five NT by PI-FGSM. \textbf{(b)}: the adversarial examples are crafted via an ensemble of the remaining five NT by DTPI-FGSM.\textbf{(c)}: the adversarial examples are crafted via an ensemble of the remaining six NT and two EAT by DTPI-FGSM.}
	\label{gamma}
\end{figure*}

\subsection{{Parameters Tuning for PIM}}
\label{tpw}
In our previous work~\cite{Gao+20}, we only discuss the non-targeted transferability of PIM, but the targeted transferability of them remains unknown. Therefore, we conduct a series of experiments to determine the appropriate parameters for targeted PIM in this section. 
For our proposed PI-FGSM and any other extensions (\textit{e.g.}, DTPI-FGSM), the amplification factor $\beta$ is the most important parameter because we cannot get excess noise without it. Hence we first analyze the effect of amplification factor $\beta$ in Sec.~\ref{af}, then specify the project factor $\gamma$ in Sec.~\ref{pf}, and finally discuss about kernel size $k_w$ of project kernel $\bm{W_p}$ in Sec.~\ref{pk}.

To conduct our ablation study, we attack different combinations of ensemble models (\textit{e.g.}, Inc-v4, Res-50, Res-101, Res-152 and IncRes-v2) and leave the remaining one model as the hold-out model (\textit{e.g.}, Inc-v3) by PI-FGSM and DTPI-FGSM to examine the targeted transferability. If the hold-out model is in EAT (\textit{e.g.}, Inc-v3$_{ens3}$), then our substitute ensemble models are the remaining six NT and two EAT, \textit{i.e.}, Inc-v3,Inc-v4, Res-50, Res-101, Res-152, IncRes-v2, Inc-v3$_{ens4}$ and IncRes$_{ens}$.

\begin{figure*}
	\centering
	\includegraphics[height=5.3cm]{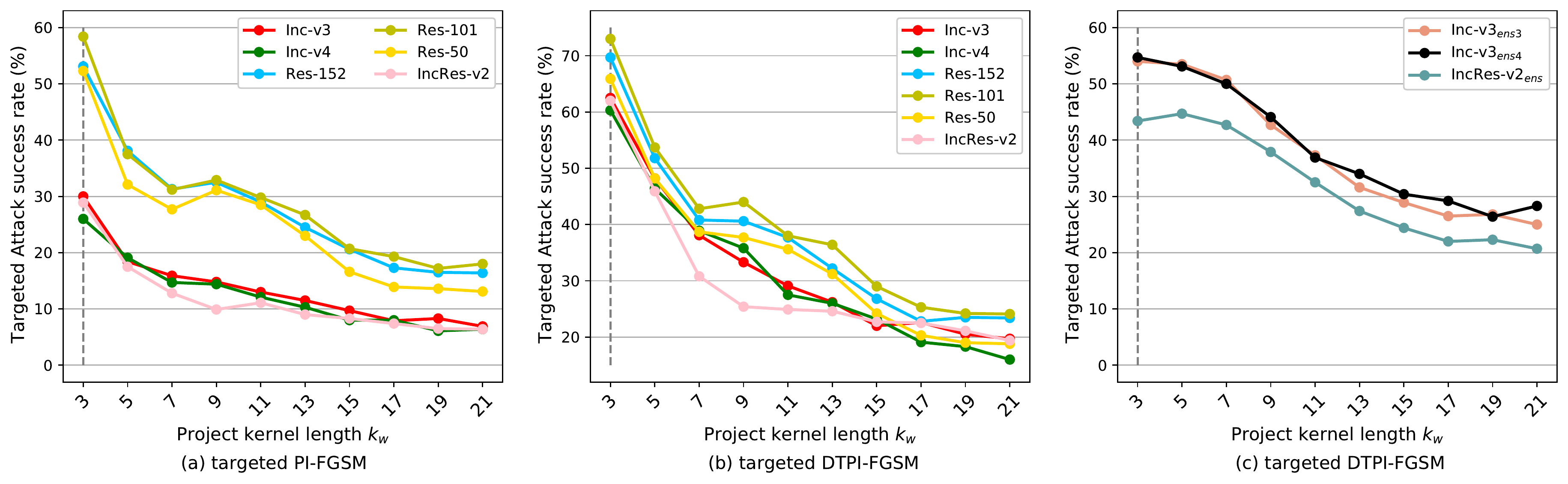}
	\caption{The targeted attack success rate(\%) of each hold-out black-box model w.r.t project kernel length $k_w$. \textbf{(a)}: the adversarial examples are crafted via an ensemble of the remaining five NT by PI-FGSM. \textbf{(b)}: the adversarial examples are crafted via an ensemble of the remaining five NT by DTPI-FGSM.\textbf{(c)}: the adversarial examples are crafted via an ensemble of the remaining six NT and two EAT by DTPI-FGSM.}
	\label{kw}
\end{figure*}
\subsubsection{The Effect of Amplification factor $\beta$}
\label{af}
In this section, we calculate the success rates of different $\beta$ which varies from 1 to the total number of iterations $T$, \textit{i.e.}, 20. The results are shown in Fig.~\ref{beta}. In general, a big amplification factor ($\textgreater 1$) will improve the performance. More specifically, any $\beta \textgreater 3$ are already boost the transferability by a large margin, especially for DTPI-FGSM.
Nevertheless, a too big amplification factor is not always better. For example, when the hold-out model is Res-50, \textit{i.e.}, Fig.~\ref{beta}(c), further increasing $\beta$ will not bring significant improvement and may even reduce the targeted transferability (the performance of $\beta=10$ is better than that of $20$). Therefore, setting a suitable amplification factor that can better escape from the local optimal of the ensemble models is essential.

Considering that the optimal $\beta$ under different models and methods are different, we set them separately:
\begin{itemize}
	\item If the attack method is PI-FGSM and the hold-out network is in NT, the optimal $\beta$ of the most substitute models are 20. So we set $\beta = 20$ in this case.
	\item If the attack method is DTPI-FGSM and the hold-out network is in NT, the optimal $\beta$ of the different substitute models are not consistent, but when the $\beta$ is greater than 10, a bigger $\beta$ has little influence on the result. So we just set $\beta = 10$ to control the step size.
\end{itemize}

\begin{figure*}[t]
	\centering
	\includegraphics[height=5.1cm]{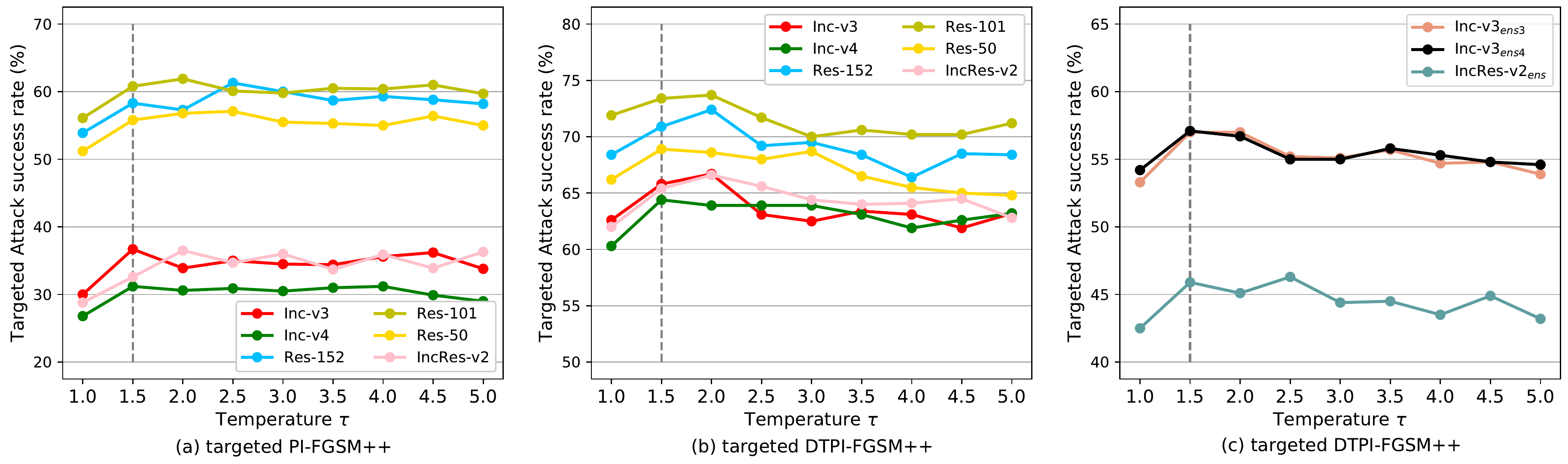}
	\caption{The targeted attack success rate(\%) of each hold-out black-box model w.r.t the temperature $\tau$. \textbf{(a)}: the adversarial examples are crafted via an ensemble of the remaining five NT by PI-FGSM++. \textbf{(b)}: the adversarial examples are crafted via an ensemble of the remaining five NT by DTPI-FGSM++.\textbf{(c)}: the adversarial examples are crafted via an ensemble of the remaining six NT and two EAT by DTPI-FGSM++.}
	\label{T}
\end{figure*}

\subsubsection{The Effect of Project Factor $\gamma$}
\label{pf} 
After determining the optimal amplification factor, we then study the effect of different project factor $\gamma$, which ranges from $0.1\beta\alpha$ to $\beta\alpha$ with the granularity $0.1\beta\alpha$.
A first glance at Fig.~\ref{gamma} shows that the performance of a bigger $\gamma$ is usually better than smaller ones. In fact, this parameter controls the influence of our proposed project term. That is to say, the bigger it is, the more significant the impact from the cut noise $\bm{C}$ is. 
From the experimental results, we can also demonstrate the effectiveness of our project term, which helps to improve the targeted transferability by a large margin. Similar to the amplification factor, the optimal $\gamma$ is also different. For example, if the attack method is PI-FGSM and the hold-out network is in NT, the choice of $\gamma=0.8\beta\alpha$ is better than any other values in most cases and increasing the $\gamma$ from $0.8$ to $1.0$ will significantly degrade the performance of our attacks. However, for DTPI-FGSM, the trends of $\gamma$ from $0.8$ to $1.0$ are stable. To avoid tuning various hyper-parameters, we fix $\gamma = 0.8\beta\alpha$ for all methods.
\eat{
	\begin{itemize}
		\item If the attack method is PI-FGSM and the hold-out network is in NT, the choice of $\gamma=0.8\beta\alpha$ is better than any other values in most cases. So we set $\gamma = 0.8\beta\alpha$.
		\item If the attack method is DTPI-FGSM and the hold-out network is in NT, the overall trend is similar to PI-FGSM. So we also set $\gamma= 0.8\beta\alpha$.
		\item If the attack method is DTPI-FGSM and the hold-out network is in EAT, the success rates of targeted attack continue increasing at first, then become stable, finally reach the peak when $\gamma = \beta\alpha$. Therefore, we set it to $\beta\alpha$ for better performance.
	\end{itemize}
}

\subsubsection{The Effect of Project Kernel Size}
\label{pk}
In fact, the size of project kernel $\bm{W_p}$ also plays a crucial role in targeted transferability. If the kernel size $k_w$ is $1\times 1$, then the project term will be invalid because the adjacent regions of the excess noise are not affected at all. Therefore, we conduct this experiment to analyze the effect of $k_w$ (the length of $\bm{W_p}$) which ranges from 3 to 21.

Here we attack different hold-out networks with the uniform kernel defined at Eq.~\ref{W_p}, and the results are shown in Fig.~\ref{kw}. Different from non-targeted PI-FGSM (see Fig. 3 of \cite{Gao+20}), a big $k_w$ usually degrade the targeted transferability. 
For example, when the hold-out network is Res-50 and the attack method is PI-FGSM, changing kernel length from 3 to 5 drastically decreases performance by about 20.2\%. Therefore, we froze the $k_w$ as 3 in our paper.

\begin{figure*}
	\centering
	\includegraphics[height=5.4cm]{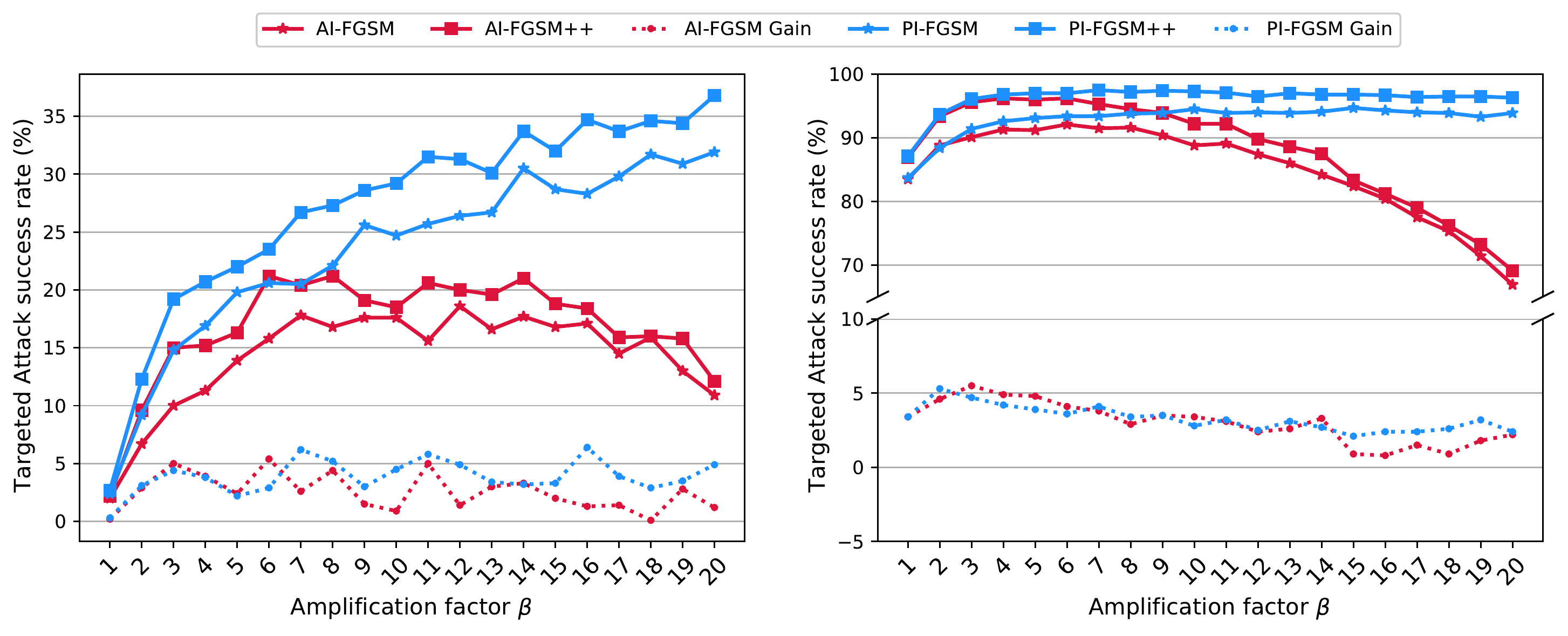}
	\caption{We analyze the promotion effect of temperature $\tau=1.5$ w.r.t amplification factor $\beta$ on targeted success rate. The adversarial examples are crafted via an ensemble of Inc-v4, IncRes-v2, Res-50, Res-101 and Res-152 by I-FGSM, I-FGSM++, PI-FGSM as well as PI-FGSM++. The hold-out black-box model is Inc-v3. 
   \textbf{Left}: the targeted success rate of hold-out Inc-v3 (Hold-out).
   \textbf{Right}: the average targeted success rate of each ensemble model (AoE). \eat{\textbf{Right}: the targeted success rate of ensemble model (Ensemble).
	} 
	From the trends of I-FGSM Gain and PI-FGSM Gain, we can demonstrate that the amplification factor and temperature $\tau$ are mutually reinforcing.
	}
	\label{ablation}
\end{figure*}

\subsection{{Parameters Tuning for PIM++}}
\label{ttpn}
Since our PIM++ bases on PIM, we follow the optimum parameters obtained from the Sec.~\ref{tpw}, which including amplification factor, project factor and kernel length.

To verify the effectiveness of PIM++, we also attack different combinations of the ensemble models (\textit{e.g.}, Inc-v4, Res-50, Res-101, Res-152 and IncRes-v2) by PI-FGSM, PI-FGSM++ and DTPI-FGSM++ and leave the remaining one model as the hold-out model (\textit{e.g.}, Inc-v3) to examine the targeted transferability. If the hold-out model is in EAT (\textit{e.g.}, Inc-v3$_{ens3}$), then our ensemble white-box models are the remaining six NT and two EAT, \textit{i.e.}, Inc-v3, Inc-v4, Res-50, Res-101, Res-152, IncRes-v2, Inc-v3$_{ens4}$ and IncRes$_{ens}$.
\begin{table}[t]
	\centering
	\caption{The adversarial examples are crafted via an ensemble of Inc-v4, IncRes-v2, Res-152, Res-101 and Res-50 by PI-FGSM and PI-FGSM++ with $\tau=1.5$. The hold-out model is Inc-v3.}
	\resizebox{0.6\linewidth}{!}{
		\begin{tabular}{c|c|c}
			\hline
			& LSM$\downarrow$   & LHM$\downarrow$ \\
			\hline
			\hline
			PI-FGSM & 62.20 & 4354.72 \\
			\hline
			PI-FGSM++ & \textbf{18.00}  & \textbf{3871.57} \\
			\hline
		\end{tabular}%
	}
	\label{lm}%
\end{table}%

\subsubsection{The Analysis of Temperature $\tau$}
Intuitively, applying any temperature $\tau \textgreater 1$ will cause an increase in the loss during the training process. Since the update goal is to decrease the loss at each iteration, a large loss can push the adversarial example toward the target label's region more efficiently. {Accompanied by the amplification factor, we argue that resultant adversarial examples will be more easily pushed to the ``global optimal region'' and gain higher transferability.}

To convincingly verify the above assumption, we propose two evaluation metrics to numerically analyze the result. One is the Loss of Substitute Model (LSM), which calculates the sum of the substitute model's loss for each adversarial example. The other is the Loss of Hold-out Model (LHM), which calculates the sum of the hold-out model's loss for each adversarial example.
To avoid cherry-picking, we compare the difference between PIM and PIM++ on all images from the dataset. As demonstrated in Tab.~\ref{lm}, with the help of soft probability distribution, our PI-FGSM++ achieves better performance on both LSM and LHM, especially for LSM. As we can see, PI-FGSM++ reduces LSM by about \textbf{80\%} compared to PI-FGSM. Therefore, we can demonstrate that our resultant adversarial examples do get close to the ``global optimal region''.


\subsubsection{The Effect of Temperature $\tau$}
In this section, we report the experimental results of our proposed PIM++ with respect to temperature $\tau$ in Fig.~\ref{T}. We tune $\tau=1.0,\,1.5\,,2.0,\,...,5.0$. Specially, when temperature $\tau=1.0$, PIM++ degrades to PIM. A first glance shows that targeted attack success rates are sensitive to the $\tau$, and a too high temperature will cause the degradation of performance, \textit{\textit{e.g.}}, the black curve in Fig.~\ref{T}(c). Besides, the optimal $\tau$ also varies from each other. For example, if the attack method is PI-FGSM++ and the hold-out network is in NT (\textit{e.g.}, Res-152), the performance of $\tau=2.5$ is often the best. However, if the attack method is DTPI-FGSM++ and the hold-out model is in EAT (\textit{e.g.}, Inc-v3$_{ens3}$), $\tau=1.5$ is better. Besides, we observe that $\tau=1.5$ already achieves great improvement for all hold-out models. To make this parameter more consistent, we set $\tau=1.5$ in our paper.
\eat{
To make this parameter more consistent, we also discuss the best settings for each of the three attack scenarios:
\begin{itemize}
	\item If the attack method is PI-FGSM and the hold-out network is in NT, we set $\tau=2.5$ because further raise the temperature cannot significantly boost the attack ability.
	\item If the attack method is DTPI-FGSM and the hold-out network is in NT, a too high temperature leads to worse transferability. So we set $\tau=1.5$.
	\item If the attack method is DTPI-FGSM and the hold-out network is in EAT, most curves reach the peak when $\tau$ is 1.5. So we set $\tau = 1.5$ finally.
\end{itemize}
}

\begin{figure}[t]
	\centering
	\includegraphics[width=0.6\linewidth]{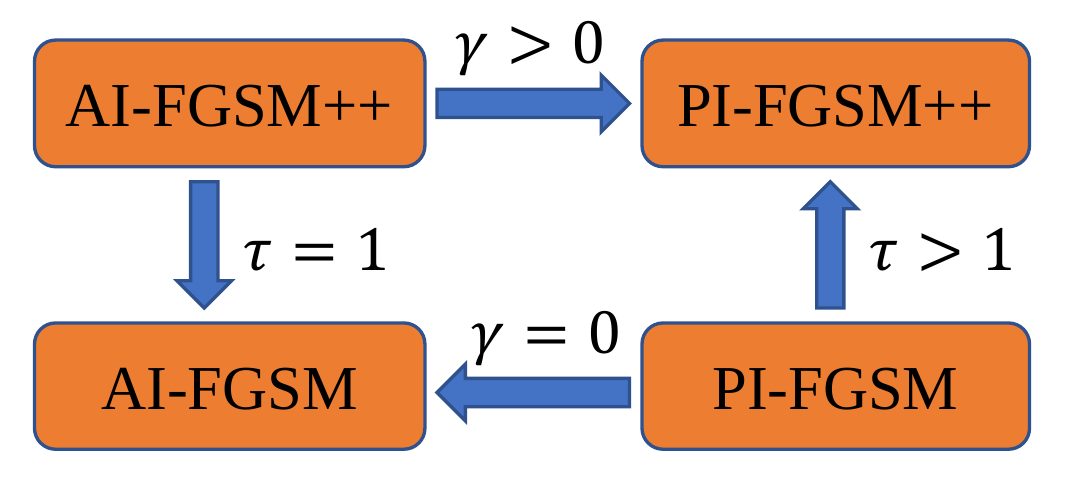}
	\caption{The relationships between AI-FGSM, AI-FGSM++, PI-FGSM and PI-FGSM++. We relate them by setting the temperature $\tau$ and project factor $\gamma$.}
	\label{relation}
\end{figure}
\eat{
\begin{figure}[t]
	\centering
	\includegraphics[height=3cm]{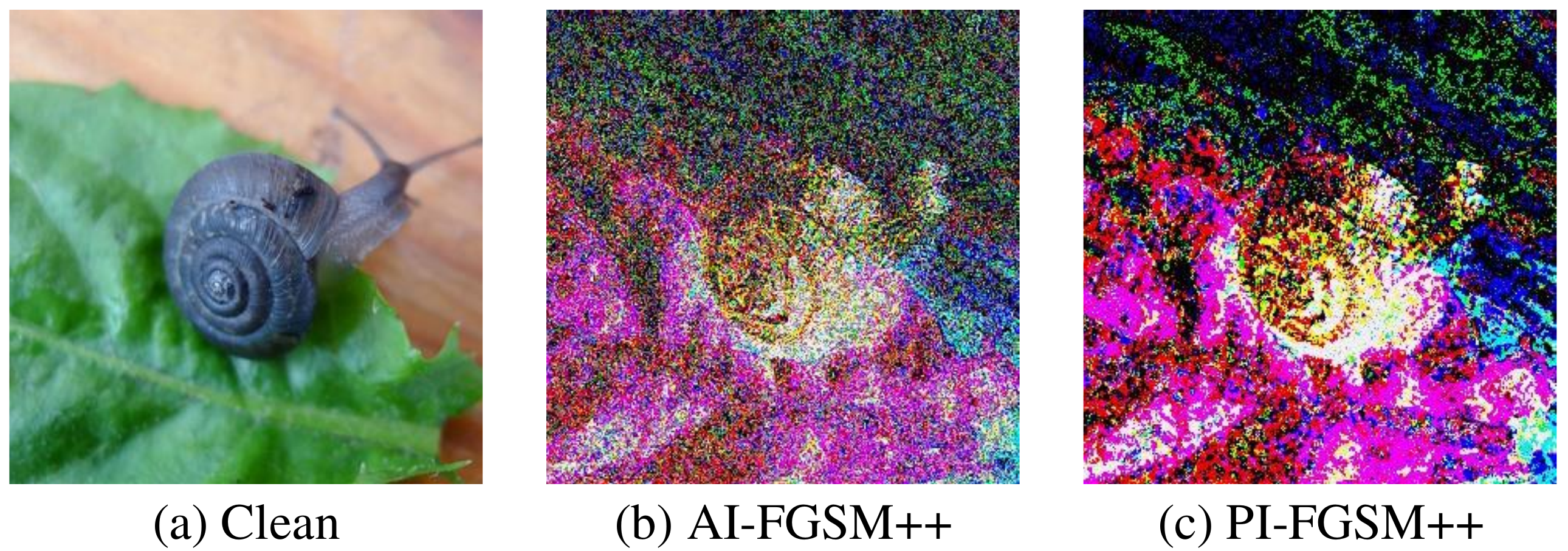}
	\caption{The visualization of (a) clean image and (b \& c) patch map generated by AI-FGSM++ and our PI-FGSM++ for Inc-v3 model~\cite{ref_article27}. The maximum perturbation $\epsilon$ is limited to 16 and the amplification factor $\beta = 10$. Our PI-FGSM++ can generate more regionally homogeneous adversarial noise than AI-FGSM++.}
	\label{pm}
\end{figure}
}

\begin{table*}[t]
	\centering
	\caption{The targeted attack success rates (\%) of FGSM, I-FGSM, TI-FGSM, TI-BIM, MI-FGSM, DI-FGSM, DTMPo-FGSM, PI-FGSM, PI-FGSM++ and DTPI-FGSM++ for an ensemble of five white-box NT (Ensemble), the average of each ensemble white-box NT (AoE) and a hold-out black-box NT (Hold-out). In each row, ``-" denotes the hold-out model and the adversarial examples are crafted via the ensemble of the remaining five NT.}
	\resizebox{0.85\linewidth}{!}{
		\begin{tabular}{c|c|c|c|c|c|c|c|c}
			\hline
			& Attack & -Inc-v3$\uparrow$ & -Inc-v4$\uparrow$& -Res152$\uparrow$ & -Res101$\uparrow$ & -Res50$\uparrow$ & -IncRes$\uparrow$ & Average$\uparrow$\\
			\hline
			\hline
			\multirow{8}[0]{*}{Ensemble (white-box)} & FGSM & 0.1  & 0.1  & 0.0  & 0.1  & 0.0  & 0.0 & 0.1 \\
			
			& TI-FGSM & 0.1  & 0.0  & 0.0  & 0.1  & 0.0  & 0.0 & 0.0 \\
			& I-FGSM & 99.8  & 99.5  & 99.2  & 99.3  & 99.3  & 99.5 & 99.4 \\
			& TI-BIM & 99.6  & 99.6  & 99.2  & 99.4  & 99.3  & 99.5 & 99.4 \\
			& MI-FGSM & 99.4  & \textbf{99.7}  & 99.7  & \textbf{99.7}  & \textbf{99.6}  & 99.7 & \textbf{99.6} \\
			& DI-FGSM & 89.7  & 91.0  & 88.1  & 86.5  & 86.2  & 90.6 & 88.7 \\
			
			& PI-FGSM (Ours) & 99.5  & 99.5  & \textbf{99.9}  & 99.0  & 99.4  & 98.7 & 99.3 \\
			& PI-FGSM++ (Ours)& \textbf{99.9}  & \textbf{99.7}  & 99.4  & 99.3  & 99.3  & \textbf{99.8} & 99.5 \\
			\cline{2-9}
			
			& DTMPo-FGSM & 88.4  & 88.9  & 88.1  & 86.4  & 85.8  & 91.1 & 88.1 \\
			& DTPI-FGSM++ (Ours)& \textbf{99.4}  & \textbf{99.2}  & \textbf{98.9}  & \textbf{98.1}  & \textbf{97.7}  & \textbf{98.8} & \textbf{98.7} \\
			\hline
			\multirow{8}[0]{*}{AoE (white-box)} & FGSM & 0.1  & 0.1  & 0.1  & 0.1  & 0.0  & 0.1 & 0.1 \\
			
			& TI-FGSM & 0.1  & 0.0  & 0.0  & 0.1  & 0.1  & 0.0 & 0.1 \\
			& I-FGSM & 83.3  & 81.4  & 79.4  & 76.2  & 78.5  & 85.2 & 80.7 \\
			& TI-BIM & 83.7  & 85.3  & 80.5  & 79.6  & 78.8  & 88.4 & 82.7 \\
			& MI-FGSM & 83.8  & 86.5  & 82.1  & 80.9  & 80.8  & 89.6 & 84.0 \\
			& DI-FGSM & 78.3  & 78.0  & 72.8  & 71.0  & 68.5  & 77.7 & 74.4 \\
			
			& PI-FGSM (Ours)& 94.1  & 94.7  & 94.7  & 93.0  & 93.1  & 93.7 & 93.9 \\
			& PI-FGSM++ (Ours)& \textbf{96.3}  & \textbf{96.6}  & \textbf{94.8}  & \textbf{94.9}  & \textbf{95.1}  & \textbf{97.0} &\textbf{95.8} \\
			\cline{2-9}
			& DTMPo-FGSM & 76.8  & 77.2  & 75.4  & 70.3  & 70.3  & 78.0 & 74.7 \\
			
			& DTPI-FGSM++ (Ours)& \textbf{97.0}  & \textbf{96.7}  & \textbf{94.8}  & \textbf{94.3}  & \textbf{94.6}  & \textbf{96.3} & \textbf{95.6} \\
			\hline
			\multirow{8}[0]{*}{Hold-out (black-box)}  & FGSM & 0.0  & 0.0  & 0.0  & 0.0  & 0.0  & 0.0 & 0.0 \\
			& TI-FGSM & 0.0  & 0.0  & 0.0  & 0.0  & 0.0  & 0.0 & 0.0 \\
			& I-FGSM & 1.8   & 2.3   & 11.2  & 11.0  & 10.0  & 2.0 &6.4 \\
			& TI-BIM & 3.2   & 1.8   & 8.7   & 9.7   & 7.7   & 1.3 & 5.4 \\
			& MI-FGSM & 6.2   & 5.6   & 19.5  & 17.9  & 14.0  & 4.1 & 11.2 \\
			& DI-FGSM & 21.1  & 19.3  & 28.9  & 27.7  & 24.5  & 17.4 & 23.2 \\
			
			& PI-FGSM (Ours)& 30.3  & 25.6  & 54.9  & 57.7  & 53.0  & 27.0 & 41.4 \\
			& PI-FGSM++ (Ours)& \textbf{36.7}  & \textbf{31.2}  & \textbf{58.3}  & \textbf{60.8}  & \textbf{55.8}  & \textbf{32.6} & \textbf{45.9} \\
			\cline{2-9}
			& DTMPo-FGSM & 37.2  & 33.6  & 39.1  & 39.5  & 36.8  & 33.9 & 36.7 \\
			
			& DTPI-FGSM++ (Ours) & \textbf{65.8}  & \textbf{64.4}  & \textbf{70.9}  & \textbf{73.4}  & \textbf{68.9}  & \textbf{65.4} &\textbf{68.1} \\
			\hline
	\end{tabular}}%
	\label{NT}%
\end{table*}%
Besides, we also conduct an ablation study to analyze the promotion effect of temperature $\tau$ with respect to amplification factor $\beta$ in Fig.~\ref{ablation}. Here we
compare the performance of AI-FGSM~\cite{Gao+20} (\textit{i.e.}, applying temperature to each iteration), AI-FGSM++ , PI-FGSM as well as PI-FGSM++, and show the trends of AI-FGSM Gain and PI-FGSM Gain (\textit{i.e.}, the results of applying temperature minus the results of its corresponding baseline).
The relationship between these four attack methods is shown in Fig.~\ref{relation} and the main difference between AI-FGSM and PI-FGSM is whether or not to add  the project term at each iteration. 

As illustrated in Fig.~\ref{ablation}, without the cooperation of amplification factor (\textit{i.e.}, $\beta=1$), the temperature term seems to have no help to improve the black-box transferability.
However, if we apply the amplification factor at each iteration, \textit{e.g.}, $\beta = 3$, we can observe that the targeted success rates of both AI-FGSM++ and PI-FGSM++ will be significantly boosted by about 5\%. It also demonstrates that the tradeoff between the amplification factor (underfitting) and the temperature term (overfitting) is effective.

\begin{table*}[htbp]
	\centering
	\caption{The targeted attack success rates (\%) of DMI-FGSM, DTI-FGSM, DTMI-FGSM, DTMPo-FGSM, DTPI-FGSM and DTPI-FGSM++ for an ensemble of eight white-box models (Ensemble), the average of each ensemble white-box models (AoE) and the hold-out black-box EAT (Hold-out). In each row, ``-" denotes the hold-out model and the adversarial examples are crafted via the ensemble of the remaining five NT and two EAT.}
	\begin{tabular}{c|c|c|c|c|c}
		\hline
		& Attack & -Inc-v3$_{ens3}$$\uparrow$  & -Inc-v3$_{ens4}$$\uparrow$  & -IncRes-v2$_{ens}$$\uparrow$ & Average$\uparrow$ \\
		\hline
		\hline
		\multirow{6}[2]{*}{Ensemble (white-box)} 
		
		& DMI-FGSM & 81.8  & 78.7  & 79.9  & 80.1  \\
		& DTI-FGSM & 54.5  & 55.3  & 57.9  & 55.9  \\
		& DTMI-FGSM & 44.3  & 46.1  & 48.4  & 46.3  \\
		& DTMPo-FGSM & 58.7  & 59.1  & 60.2  & 59.3  \\
		& DTPI-FGSM (ours) & 93.3  & 93.4  & 94.0 &  93.6 \\
		& DTPI-FGSM++ (ours) & \textbf{93.4}  & \textbf{93.9}  & \textbf{94.4}  & \textbf{93.9}  \\
		\hline
		\multirow{6}[2]{*}{AoE (white-box)} 
		
		& DMI-FGSM & 63.5  & 61.3  & 62.1  & 62.3  \\
		& DTI-FGSM & 43.7  & 44.5  & 46.0  & 44.7  \\
		& DTMI-FGSM & 35.7  & 36.2  & 37.9  & 36.6  \\
		& DTMPo-FGSM & 48.6  & 48.7  & 50.8  & 49.4  \\
		& DTPI-FGSM (ours) & 86.1  & 86.8  & 87.6  & 86.8  \\
		& DTPI-FGSM++ (ours) & \textbf{87.9}  & \textbf{88.1}  & \textbf{88.7}  & \textbf{88.2}  \\
		\hline
		\multirow{6}[2]{*}{Hold-out (black-box)} 
		
		& DMI-FGSM & 0.9   & 1.3   & 0.9   & 1.0  \\
		& DTI-FGSM & 12.3   & 11.7   & 10.4   & 11.5  \\
		& DTMI-FGSM & 13.4  & 14.4  & 13.8  & 13.9  \\
		& DTMPo-FGSM & 21.5  & 20.3  & 19.1  & 20.3  \\
		& DTPI-FGSM (ours)& 54.0  & 54.7 & 42.5  &  50.4 \\
		& DTPI-FGSM++ (ours)& \textbf{57.0}  & \textbf{57.1}  & \textbf{46.0}  & \textbf{53.4}  \\
		\hline
	\end{tabular}%
	\label{EAT}%
\end{table*}%

\subsection{Comparison on NT}
\label{attacking nt}
After the above discussion, we further compare our PI-FGSM, PI-FGSM++ and DTPI-FGSM++ with state-of-the-art targeted attacks, including FGSM, I-FGSM, MI-FGSM, DI-FGSM, TI-FGSM, TI-BIM, DTMPo-FGSM. The success rates of targeted attacks are reported in Tab.~\ref{NT}.  Our adversarial examples are crafted via an ensemble of five white-box NT. Then we evaluate the attack success rates on these ensemble models (white-box attack) and the remaining hold-out model (black-box attack). As for the ensemble strategy, we show at Eq.~\ref{ensemble}. Specially, we set the weight of each ensemble model to be $1/5$ in this case.

We first analyze the performance of the black-box attack. 
It can be observed that our targeted PI-FGSM already surpasses the transferability of state-of-the-art DTMPo-FGSM, which is integrated with several well-known approaches including DI-FGSM, MI-FGSM, TI-BIM and Po-FGSM. Besides, our PI-FGSM++ further improves the transferability by an extra \textbf{4.5\%}, and it also demonstrates the effectiveness of temperature term. Furthermore, our proposed DTPI-FGSM++ can make the targeted attack success rate more impressive.
Remarkably, the adversarial examples crafted by it can reach \textbf{68.1\% }success rate on average which surpasses the DTMPo-FGSM by \textbf{31.4\%}. Besides, when the hold-out network is Res-101, the margin can be further enlarged to \textbf{33.9\%}.

Although our method has excellent black-box attack ability, it does not sacrifice performance on the white-box model. 
As demonstrated in Tab.~\ref{NT}, all variants of our method achieve a success rate of nearly \textbf{100\%} in the white-box (Ensemble) case. By contrast, the performance of DI-FGSM and DTMPo-FGSM is lower than 90\%. To have a better comparison, we also evaluate the success rates in the AoE case. A first glance shows that the attack success rates of all other approaches are sharply degenerate, \textit{e.g.}, the average success rate of I-FGSM drops from 99.4\% (Ensemble case) to 80.7\%. However, our approaches still maintain a strong white-box attack ability, \textit{i.e.}, all over \textbf{93.0\%}. This phenomenon may be attributed to the amplification factor and the project term which help jump out of the local optimum more easily, thus not hindering performance on each white-box model significantly. Compared with PI-FGSM, our PI-FGSM++ can further push the resultant adversarial examples toward ``global optimal region'', \textit{e.g.}, improving the average success rate of AoE by an extra \textbf{1.9\%}.

\subsection{Comparison on EAT}
\label{attacking eat}
Adversarial training is currently recognized as the most effective way to resist strong attacks. When the adversarial examples transfer to the adversarially trained models, the attack success rate is usually very low, especially for targeted attacks. Therefore, crafting transferable targeted adversarial examples for black-box adversarially trained models is more challenging than normally trained models.

To better demonstrate the performance of our method, here we conduct another experiment to verify the effect of our proposed method. The adversarial examples are generated via an ensemble of Inc-v3, Inc-v4, Res-50, Res-101, Res-152, IncRes-v2 and any two models from Inc-v3$_{ens3}$, Inc-v3$_{ens4}$ and IncRes-v2$_{ens}$ by DMI-FGSM, DTI-FGSM, DTMI-FGSM, DTMPo-FGSM, DTPI-FGSM and DTPI-FGSM++, and the remaining one model in EAT serves as the hold-out model to test the transferability of the above attack methods.

As shown in Tab.~\ref{EAT}, our method is superior to other approaches by a large margin. In the black-box setting, compared with DTMI-FGSM and DTMPo-FGSM whose success rates are 13.9\% and 20.3\% on average, our DTPI-FGSM++ has \textbf{53.4\%} success rate against EAT which improves by \textbf{39.5\%} and \textbf{33.1\%} respectively. Specially, when the hold-out network is Inc-v3$_{ens4}$, DTPI-FGSM++ can outperform DTMPo-FGSM by \textbf{36.8\%} at most. Besides, we observe that success rates of the white-box case usually boost effectively whatever the substitute models are. For example, when applying DTPI-FGSM++, the resultant adversarial examples achieve the average success rates of \textbf{93.4\%} and \textbf{88.2\%} in Ensemble and AoE cases. However, DTI-FGSM only gets 55.9\% and 44.7\% success rates on these two evaluation metrics.  

The above results demonstrate the effectiveness of our proposed method. It also raised the real security issues that current defense methods are still vulnerable and cannot be safely deployed in real-world applications.

\section{Conclusion}
In this paper, patch-wise and patch-wise++ iterative methods are proposed to improve the transferability of adversarial examples without significantly sacrificing the performance of the white-box attack. The patch-wise iterative method (PIM), which introduces an amplification factor to the step size at each iteration and a project kernel properly assigns one pixel's overall gradient over
owing the $\epsilon$-constraint to its surrounding regions, effectively boosts both the non-targeted and targeted attacks. However, the amplification factor may hinder the further improvement of targeted transferability. To address this issue, we integrate the temperature term into our PIM and propose a PIM++ -- a black-box targeted attack towards mainstream normally trained and defense models. Our method can be generally integrated to any gradient-based attack methods. In this way, without significantly sacrificing the performance of the white-box attack, our adversarial examples can have strong transferability. Compared with the current state-of-the-art attacks, we significantly improve the success rate of targeted attacks by 33.1\% for defense models and 31.4\% for normally trained models on average.


%



\ifCLASSOPTIONcompsoc
\section*{Acknowledgments}

\else
\section*{Acknowledgment}
\fi
This work is supported by the Fundamental Research Funds for the Central Universities (Grant No. ZYGX2019J073), the National Natural Science Foundation of China (Grant No. 61772116, No. 61872064, No.61632007, No. 61602049), The Open Project of Zhejiang Lab (Grant No.2019KD0AB05).

\ifCLASSOPTIONcaptionsoff
  \newpage
\fi



%




\bibliographystyle{IEEEtran}
\bibliography{egbib}

\end{document}